\documentclass[twocolumn, switch]{article} % Method A for two-column formatting

\usepackage{preprint}

\usepackage{amsmath, amsthm, amssymb, amsfonts}

\usepackage[numbers,square]{natbib}
\usepackage[utf8]{inputenc}	% allow utf-8 input
\usepackage[T1]{fontenc}	% use 8-bit T1 fonts
\usepackage{xcolor}		% colors for hyperlinks
\usepackage[colorlinks = true,
            linkcolor = purple,
            urlcolor  = blue,
            citecolor = cyan,
            anchorcolor = black]{hyperref}	% Color links to references, figures, etc.
\usepackage{booktabs} 		% professional-quality tables
\usepackage{nicefrac}		% compact symbols for 1/2, etc.
\usepackage{microtype}		% microtypography
\usepackage{lineno}		% Line numbers
\usepackage{float}			% Allows for figures within multicol

\usepackage{lipsum}		%  Filler text

\usepackage{listings} % for computer coding

\usepackage{newfloat}
\DeclareFloatingEnvironment[name={Supplementary Figure}]{suppfigure}
\usepackage{sidecap}
\sidecaptionvpos{figure}{c}

\usepackage{titlesec}
\titlespacing\section{0pt}{12pt plus 3pt minus 3pt}{1pt plus 1pt minus 1pt}
\titlespacing\subsection{0pt}{10pt plus 3pt minus 3pt}{1pt plus 1pt minus 1pt}
\titlespacing\subsubsection{0pt}{8pt plus 3pt minus 3pt}{1pt plus 1pt minus 1pt}

\title{QRMine: A python package for triangulation in Grounded Theory}

\usepackage{xwatermark}
\newwatermark[firstpage,color=gray!90,angle=0,scale=0.28, xpos=0in,ypos=-5in]{*correspondence: \texttt{eapenbp@mcmaster.ca}}

\usepackage{authblk}

\author[1]{Bell Raj Eapen}
\author[1]{Norm Archer}
\author[2]{Kamran Sartipi}
\affil[1]{Information Systems, McMaster University, Hamilton, ON, Canada}
\affil[2]{Computer Science, East Carolina University, Greenville, NC, USA}

\begin{document}

\twocolumn[ % Method A for two-column formatting
  \begin{@twocolumnfalse} % Method A for two-column formatting
  
\maketitle

\begin{abstract}

Grounded theory (GT) is a qualitative research method for building theory grounded in data. GT uses textual and numeric data and follows various stages of coding or tagging data for sense-making, such as open coding and selective coding.  Machine Learning (ML) techniques, including natural language processing (NLP), can assist the researchers in the coding process. Triangulation is the process of combining various types of data. ML can facilitate deriving insights from numerical data for corroborating findings from the textual interview transcripts. We present an open-source python package (QRMine) that encapsulates various ML and NLP libraries to support coding and triangulation in GT. QRMine enables researchers to use these methods on their data with minimal effort. Researchers can install QRMine from the python package index (PyPI) and can contribute to its development. We believe that the concept of 'computational triangulation' will make GT relevant in the realm of big data.

\end{abstract}
%\keywords{First keyword \and Second keyword \and More} % (optional)
\vspace{0.35cm}

  \end{@twocolumnfalse} % Method A for two-column formatting
] % Method A for two-column formatting

\section{Introduction}

Qualitative Research (QR) is undergoing a paradigmatic transformation with the increasing popularity of big data, machine learning and artificial intelligence \cite{wiedemann2013opening}. Qualitative research methods such as Grounded Theory (GT), though heavily reliant on data, have an predominantly interpretive and subjective world view \cite{glaser1967discovery}. The subjectivity in QR is its strength, although researchers from the quantitative domain see it as a weakness. Qualitative researchers have traditionally eschewed the computational techniques and tools that are used for objective data analysis.

Natural language processing (NLP) and some of the numerical machine learning (ML) techniques can give insights on qualitative and quantitative data to researchers without being too intrusive into the philosophical assumptions of interpretivism. We call this process \emph{computational triangulation} (CT). these techniques may be technically challenging for social science researchers to use without a background in computer programming. In this article, we introduce QRMine \cite{eapenbr2019qrmine} (pronounced Karmine), a python package that helps to reduce this technical barrier. The theoretical basis of CT will be discussed elsewhere.

QRMine is an open-source python package that wraps some of the popular NLP and ML libraries into an easy to use command-line tool. QRMine aligns with the philosophical assumptions and the traditional stages of coding in GT. The numerical ML techniques help in the computational triangulation of numerical data to corroborate emergent insights from textual qualitative data.

\section{Grounded Theory}

GT is a qualitative research method with an emphasis on “generating theory grounded in data that has been systematically collected and analysed” \cite{strauss1990basics}. Some of the characteristics of GT that differentiates it from other qualitative methods are constant comparison by simultaneous collection and analysis of data, theoretical sampling \cite{glaser1970theoretical} based on the emergent theory, and its emphasis on the ‘theoretical sensitivity’ \cite{glaser1978advances2} or the insight of the researcher. The aim of GT is to converge towards a theory that adequately explains the phenomenon under study.

GT has been gaining traction due to the recent emphasis on data. However, researchers from the quantitative domain may consider GT chaotic and regard the theories that emerge from such studies as having limited validity, reliability and credibility \cite{wasserman2009problematics}. The interpretive and subjective nature of GT makes analysis non-reproducible which is considered by some researchers as its weakness. 

\subsection{Coding in GT}

Coding is the process of tagging various types of data, to define what each segment of the data is about, to help in sensemaking and to converge towards analytic interpretations. \cite{charmaz2006constructing} Coding is challenging for new researchers as it requires a lot of skill and experience to identify core concepts in the data and to identify their relationships with each other. 

Coding in GT comprises several stages that may vary according to the specific GT tradition followed by the researcher \cite{walker2006grounded}. In the open coding stage, data, often in the form of textual interview transcripts, are analyzed line-by-line to identify commonly occurring concepts or categories. Researchers are expected to employ a ‘constant comparison’ process during this step. Axial coding is the process of identifying relationships among the open codes in the form of ‘properties’ and ‘dimensions’ \cite{walker2006grounded}. Selective coding identifies core categories that can represent a group of codes. Natural language processing (NLP) can assist the researcher in the coding process, especially when using a large corpus of text \cite{nelson2017computational}. 

\section{Natural Language Processing (NLP) for Coding}

NLP is a field of ML and Artificial Intelligence (AI) concerned with the process of analyzing and interpreting large amounts of natural language data. Moderm NLP relies on ML algorithms and probabilistic models. Some of the commonly used tasks in NLP can be directly applied to coding in GT. Lemmatization and named entity recognition (NER) can be used to identify and count commonly occuring concepts in a large corpus of text \cite{nadeau2007survey}. This makes NER useful for identifying GT categories during the open coding stage.

Parts of speech tagging \cite{voutilainen2003part}can be used to identify subject-verb-object triads, a useful technique for supporting axial coding. Topic modelling is useful during the stage of selective coding to identify representative concepts for a set of categories. A structured methodology for applying the various NLP methods for coding in grounded theory is called computational grounded theory (CGT) \cite{nelson2017computational}. GT triangulates many other types of data in addition to text for theory building.

\section{Triangulation}

Triangulation refers to the use of multiple methods and data sources in qualitative research to develop a comprehensive understanding of phenomena \cite{patton1990sampling}. Though triangulation can involve multiple investigators, methods, theories and sources, we emphasis the simultaneous triangulation of qualitative and quantitative data which is the most challenging. The aim of triangulation is to corroborate findings from one stream of data with others. However, there is no universally accepted methodology for combining data streams \cite{jick1979mixing}.

Though a mixed-method multi-paradigm view is needed for combining numerical data with qualitative data, we posit that machine learning methods can be instrumental in including numerical data, especially big data in qualitative enquiries without epistemological contradictions. QRMine applies NLP and ML on both textual and numerical data helping researchers to derive complementary insights from both.

\section{Machine Learning (ML) for Triangulation}

Generally, the triangulation of insights from numerical data is based on inferential statistics.  Inferential statistical methods have a positivist ontology that is difficult to reconcile with a qualitative study without making it a multi-paradigmatic mixed-method study. Machine learning techniques (although numerical at the core) can be non-deterministic with a subjective interpretive worldview that aligns with qualitative research. 

Another challenge in triangulation is the volume of big data and the sparsity of multidimensional data that make inferences difficult to deduce. The curse of dimensionality \cite{donoho2000high} makes conventional statistical tests irrelevant in such situations. ML techniques can be useful in these circumstances to derive insights.

Next, we describe the design and usage of QRMine.

\section{QRMine}

QRMine is an open-source python package that provides an easy to use wrapper around NLP and ML packages. QRMine makes coding of textual data and deriving insights from numerical data less challenging for non-technical researchers. 

\subsection{Design and Testing}
    
The text analysis functions use the textacy \cite{dewilde2017textacy} and Spacy \cite{honnibal2015spacy} python packages for tokenizing and part-of-speech tagging. In GT, it is recommended to 'Code for action' during the open coding phase \cite{charmaz2006constructing}. Hence QRMine treats repeating verbs as the categories in open coding. The coding dictionary is created by identifying adjacent concepts (properties) and their adjectives and adverbs (dimensions). 

Some of the other packages used by QRMine are as follows: VaderSentiment is used for the sentiment analysis from text \cite{chauhan2018twitter}. The ML functions are implemented using Keras and TensorFlow \cite{srinivasa2018natural}.  'Click' is used to implement command-line inputs and for formatting outputs \cite{click4qrmine}.  Other packages used are imbalanced-learn for oversampling rare events, and mlxtend \cite{raschka2018mlxtend}.

Appropriate defaults are set for most parameters. Users can use command-line options to perform most of the analysis. The output is useful to derive insights from data. QRMine does not support the interactive coding of data. Many commercial software packages such as NVivo and  Dedoose \cite{lieber2013dedoose} are available if interactive coding is needed.

QRMine is not yet tested on real data. We hope that the research community will use it and report issues and help us collaboratively to develop this open-source tool.

\subsection{Availability}

QRMine is hosted at https://github.com/dermatologist/nlp-qrmine. The package is available from the python package index (PyPI.org) and can be installed using pip [see below].
\begin{lstlisting}
pip install qrmine
\end{lstlisting}
QRMine depends on the spacy English language model that is not available on pypi. This can be installed after the previous step as shown below:
\begin{lstlisting}
python -m spacy download en_core_web_sm
\end{lstlisting}
The various modules of QRMine package can be imported into any other python code or jupyter notebook as below.

\begin{lstlisting}
from qrmine import ReadData
from qrmine import Qrmine
from qrmine import MLQRMine
\end{lstlisting}

The ReadData module imports functions for importing data, Qrmine implements NLP and MLQRMine implements the ML functions. The modules are described in the documentation \cite{eapenbr2019qrmine}.

\subsection{Importing Data}
    
The text input file is specified by the -i flag. Multiple input files can be supplied at the command-line with repeating -i flags. The results are directed to the \emph{stdout} by default, but can be sent to a file using the -o flag.

\begin{lstlisting}
qrmine  -i transcript.txt \
        -o output.txt
\end{lstlisting}

QRMine can import individual documents or interview transcripts in a single text file with a ‘<break>TOPIC</break>’ tag separating topics or sections [see example below]. Multiple tags of the same type are supported. This is useful when the transcript includes a conversation with many participants as follows: 
    
\begin{verbatim}
First interview with John.
Any number of lines with
the transcribed text
<break>Interview_John</break>

Second interview with Jane.
More text.
<break>Interview_Jane</break>

Additional comments by John.
Shows that the tag can be repeated.
<break>Interview_John</break>
\end{verbatim}

Numeric data is supplied as a single csv file with the identifier as the first column, followed by independent variable and the dependent variable as the last column. The identifier in the first column can be text and can be used to link to documents or transcripts while all other columns should be numeric
[See example below].

\begin{verbatim}
index, obesity, bmi, exercise, fbs, has_diabetes
1, 0, 29, 1, 89, 1
2, 1, 32, 0, 92, 0
......

\end{verbatim}
    \subsection{Data analysis with QRMine}
    
The package can be installed from the python repository with \emph{pip install qrmine}. The word vectors for spacy have to be separately installed \cite{spacyvector}. QRMine installs as a command line script and can be invoked directly. All command-line flags are documented \cite{eapenbr2019qrmine}. 

For example, the following commands show the top 10 categories (for open coding) and generate the coding dictionary (for axial coding).

\begin{lstlisting}
 qrmine -i transcript.txt \
        --cat \
        --codedict -n 10 
\end{lstlisting}

The following command lists the top three topics and assigns the documents to these topics.

\begin{lstlisting}
 qrmine -i transcript.txt \
        --topics \
        --assign -n 3 
\end{lstlisting}

The coding dictionary, topics and topic assignments can be created from the entire corpus using the respective command line options. Categories (concepts), summary and sentiment can be viewed for the entire corpus or particular topics specified using the --titles flag. Sentence level sentiment output is possible with the --sentence flag. Filtering documents based on sentiment, titles or categories is possible for further analysis, using --filters or -f option. 

The script below demonstrates how to find the sentiment of two segments of the text with the titles P5 and P7.

\begin{lstlisting}
qrmine  -i transcript.txt \
        -t P5 -t P7 \
        --sentiment 
\end{lstlisting}

This is how a coding dictionary can be generated from documents having a positive sentiment.

\begin{lstlisting}
qrmine  -i transcript.txt \
        -f pos --codedict 
\end{lstlisting}

The numeric function supported are neural network (--nnet) that displays the accuracy of the model after a certain number of epochs; support vector machine (--svm) producing the confusion matrix as output; kmeans clustering (--kmeans) showing the cluster assignments; k-nearest neighbours (-knn) displaying k nearest rows to a specified row; and pca (--pca) showing the factors. Many of the ML functions such as neural network take a second argument (-n) as shown below. 
\begin{lstlisting}
qrmine  -i --csv data.csv \
        --nnet -n 10
\end{lstlisting}

The -n argument represents the number of epochs in nnet, the number of clusters in kmeans, the number of factors in pca, and the number of neighbours in KNN. KNN also takes the --rec or -r argument to specify the record [see below]. 
\begin{lstlisting}
qrmine  -i --csv data.csv \
        --knn -n 5 -r 733
\end{lstlisting}
Variables from csv can be selected using --titles (defaults to all). 
\begin{lstlisting}
qrmine  -i --csv  diabetes.csv \
        -t Index -t Exercise -t Obesity \
        -t Stress -t Outcome --nnet
\end{lstlisting}
The first variable will be ignored (index) and the last should be the dependent variable (DV).

\section{Limitations and future plans}

QRMine does not currently support graphical visualization of results, which may be informative in analyses such as KNN. The command-line interface with its various options may be daunting for new users. Converting transcript files to a text only format with topic tags may be time consuming. QRMine does not support data cleaning and may be sensitive to missing data. Methods for explicitly connecting qualitative codes to quantitative data points have not been implemented yet.

In the future, we expect to build visualization methods for displaying the relationship between concepts. A simple user interface may be useful. We plan to provide additional methods for qualitative to quantitative linking and for finding association rules.

\section{Discussion and Conclusion}

Machine learning is traditionally seen as an extension of statistics with a quantitative mindset. ML is useful in inductive research to derive insights from textual and numerical data. It also provides tools for combining insights from multi-modal data which is important for theory building in areas such as social sciences.

Computational Grounded Theory (CGT) is a concept introduced to leverage NLP for coding in GT research \cite{nelson2017computational}. QRMine extend CGT into computational triangulation (CT) utilizing NLP and ML techniques for linking textual and numeric data, following the traditional coding methods of GT.

The lack of tools to facilitate usage of NLP and ML for researchers is a challenge that impacts the wider acceptance of these techniques. We provide an open-source python package that can be used in the context of jupyter notebooks \cite{kluyver2016jupyter} to analyze data without much coding. We hope that we can improve on the current version, by introducing more functions that augment researchers’ abilities to derive insights from data. We urge the open-source community to contribute to the code and to potential users to report issues on our repository \cite{eapenbr2019qrmine} so that we can fix them.

%%%%%%%%%%%%%%   Bibliography   %%%%%%%%%%%%%%
\normalsize
\bibliography{qrmine}

%%%%%%%%%%%%  Supplementary Figures  %%%%%%%%%%%%
%\clearpage

%%%%%%%%%%%%%%%%   End   %%%%%%%%%%%%%%%%

\end{document}